\documentclass[journal]{IEEEtran}

\usepackage{bm} 
\usepackage{bbm}
\usepackage{tikz}
\usepackage{ifthen}
\usepackage{standalone}
\usepackage{verbatim}
\usepackage{amsfonts}
\usepackage{cite}
\usepackage{graphicx}
\usepackage{amsmath}
\usepackage{array}
\usepackage{epstopdf}
\usepackage{caption}
\usepackage{subcaption}
\usepackage{standalone}
\usepackage{enumitem}
\usepackage{booktabs}
\usepackage[ruled]{algorithm2e}
\usepackage{url}
\usepackage[subtle]{savetrees}

\usepackage{txfonts}

\begin{document}
\title{HABNet: Machine Learning, Remote Sensing Based Detection and Prediction of Harmful Algal Blooms}
\author{P.R.~Hill,~\IEEEmembership{\,\,Member,~IEEE,}, A. Kumar, M. Temimi and D.R. Bull,\IEEEmembership{\,\,Fellow,~IEEE,}

\thanks{P.R. Hill and D.R. Bull are with the Department
of Electrical and Electronic Engineering, The University of Bristol, BS8 1UB, UK. e-mail: (see http://www.bristol.ac.uk/csprh.html).}
\thanks{M. Temimi and A. Kumar are with Khalifa University: KUSTAR, UAE}
\thanks{This work was funded by The British Council: Award ref 279334808}}
\markboth{JOURNAL NAME}%
{Hill \MakeLowercase{\textit{et al.}}: HABNet: Machine Learning, Remote Sensing Based Detection and Prediction of Harmful Algal Blooms}

\maketitle
\begin{abstract}

This paper describes the application of machine learning techniques to develop a
state-of-the-art detection and prediction system for spatiotemporal events
found within remote sensing data; specifically, Harmful Algal Bloom events
(HABs).   We propose an HAB detection system based on: a ground truth historical record of HAB events, a
novel spatiotemporal datacube representation of each event (from MODIS and GEBCO bathymetry data) and a variety of machine learning
architectures utilising state-of-the-art spatial and temporal analysis methods
based on Convolutional Neural Networks (CNNs), Long Short-Term Memory (LSTM)
components together with Random Forest and Support Vector Machine (SVM) classification
methods. 

This work has focused specifically on the case study of the detection of Karenia
Brevis Algae (K. brevis) HAB events within the coastal waters of Florida (over 2850 events
from 2003 to 2018; an order of magnitude larger than any
previous machine learning detection study into HAB events).

The development of multimodal spatiotemporal datacube data structures and
associated novel machine learning methods give a unique architecture for the
automatic detection of environmental events. Specifically, when applied to the
detection of HAB events it gives a maximum detection accuracy of 91\% and a Kappa
coefficient of 0.81 for the Florida data considered.

A HAB forecast system was also developed where a temporal subset of each datacube was
used to predict the presence of a HAB in the future.  This system was not
significantly less accurate than the detection system being able to predict with
86\% accuracy up to 8 days in the future.

\end{abstract}






\section{INTRODUCTION}
\label{sec:intro} 

\noindent Algal Blooms are defined as high concentrations of
phytoplankton (algae).  Harmful Algal Blooms (HABs) are problematic algal blooms
causing toxicity and associated environmental impacts.  Often termed ``Red
Tides'', HABs have been a significant world-wide research topic over three
decades \cite{Steidiner81, blondeau2014review, stumpf2003monitoring,
tomlinson2004evaluation, gokaraju2011, gokaraju2012ensemble,song2015learning}.

They continue to be of major concern, not only due to their considerable
environmental and societal impact but also a recent significant increase in frequency reported
around the world~\cite{blondeau2014review}.

HABs can cause severe environmental and human health problems together with
associated economic impacts. Environmental impacts include mass fish stock and
marine wildlife kills. Human impacts include toxic reactions to affected seafood
and in extreme cases, fatalities.  Economic impacts include adverse effects on
beach and coastal tourism based activities together with impacts on coastal
based industries (e.g.\ fishing).  Within the United
States alone, HABs cause an estimated annual economic loss of at least \$82
million \cite{Hoagland}.

Many factors have been cited as causes of HABs but are generally caused by
favourable environmental conditions, including increasing nutrient levels
\cite{Santoleri2003}, light availability \cite{Gohin2003}  water column stratification and/or changes in water temperature~\cite{Thomas2003}.

Conventionally, the measuring of algae concentrations has relied on direct water
sampling for lab-based cell taxonomy.  These manual methods of detection and
analysis are extremely labour intensive and are limited spatially and
temporally~\cite{Craig2006}.  Conversely, remote sensing based detection methods have excellent
coverage in time and space and offer analysis systems that are not
labour intensive.  However, remote sensing based detection methods often rely on
estimated remote sensing products such as Chlorophyll: Chl-a that themselves may be
unreliable estimates and not a direct
measurement (and therefore not precisely accurate) of cell concentrations.

 HABs have a spatiotemporal footprint that ranges from weeks to months and from
 a few square kilometres to thousands of square kilometres
 \cite{mcclain2009decade,blondeau2014review}.  It is implicit that these HABs
 are spatially and temporally dependent and for the most effective detection and
 prediction a combined spatial and temporal analysis is required. 

\subsection{Background and Contributions}

\noindent HAB monitoring and forecasting using remote sensing data was first proposed
by Steidiner and Haddad in 1981~\cite{Steidiner81} utilising data from the Coastal
Zone Color Scanner (CZCS) sensor onboard Nimbus-7, operational
during the 1970s and 1980s.

This work subsequently led to a large number of remote sensing detection,
monitoring and forecasting systems developed for more recent sensors and
satellites such as MODIS-Aqua, MODIS-Terra, SeaWiFS, MERIS and more recently
Sentinel-3 \cite{blondeau2014review}.  The methods used for detection,
monitoring and forecasting of HAB events have included: reflectance band-ratio
based detection; reflectance classification (using anomaly detection); satellite
product based detection (using thresholds etc.); and spectral band differences.

The most successful and important methods for HAB detection have used spectrally
derived products such as Chl-a (Chlorophyll concentration estimate), as phytoplankton increases the
backscattered light within pigment absorption spectral frequencies.  An
excellent review of these historical and current methods, sensors and satellites
is given by Blondeau-Patissier et al.  \cite{blondeau2014review}.

There is currently no nationwide or international HAB forecasting system for HABs.
However, there are specific areas covered by HAB forecasting systems such as
NOAA's HAB forecasting systems (HAB Operational Forecast System: HAB-OFS)
for the Florida region \cite{HABOFS}.  However, this system only forecasts up to
4 days, focuses mainly on the human impact of HABs (respiratory effects etc.)
and does not use trained machine learning, capable of generating the most
effective predictions.  The HABS Observing System (HABSOS) is a detailed
observation system of HABs within the Gulf of Mexico has also been developed by
NOAA \cite{HABSOS}.  HAB-OFS and HABSOS provide forecasts to stakeholders such as local
resource and environmental managers, the seafood industry and those managing tourism
activities.

Within this paper, state-of-the-art supervised machine learning systems are proposed
for HAB detection and prediction within the region of the Gulf of Mexico and
also in an alternative case study within the Arabian Gulf. 

We conjecture that large scale spatial patterns play an extremely
significant role in the effective detection and prediction of HABs.  We have
therefore utilise machine learning tools that not only effectively characterise spatial
patterns but combine them with time series analysis machine learning tools such
as LSTMs.

\subsection{Contributions}

\noindent The contributions of this paper are as follows:

\begin{itemize}[leftmargin=*]
\item The definition, pre-processing and analysis of a large ground truth database of positive and
	negative HAB events
\item The creation of a flexible ``datacube'' supervised training structure for
	machine learning detection and forecasting of HABs
\item The demonstration of the use of state of the art machine learning
	techniques to generate optimal detection and prediction performance of
	HABs using the datacube supervised learning structure
	\begin{itemize}
		\item The evaluation of state-of-the-art machine learning techniques
		including a range of deep network architectures and topologies.
	\end{itemize}
\item The extraction of a range of features from satellite modalities.  Additionally, the
	the performance of the features in terms of their contribution to the correct classification of HAB events is ranked and the feature ranks analysed.
\item The analysis of the forecasting ability of the system using a varying number of
	days into the future (see Fig.~\ref{fig:datacube40})
\item The analysis of the ability of the system to detect HABs for a varying number
	days in the future (see Fig.~\ref{fig:datacube40})
\item Development of a highly effective and efficient HAB detection and prediction system that could be integrated within a GIS system for flexible detection, prediction and visualisation of HABs 
\end{itemize}

\noindent This paper is organised as follows.  Section II gives an
overview of the problem and applicable machine learning systems. Section
\ref{sec:HAB1} gives an overview of the proposed HAB detection and prediction
methods.  Section \ref{sec:Datacubes} describes the ``datacube'' datastructure
and the creation of a large number of datacubes from the ground truth database.
Section \ref{sec:preDC} then describes the pre-processing of the datacubes in
order to make them ready to be ingested into the machine learning system.  Then
section \ref{sec:MLS} describes and illustrates the created machine learning
structure and section \ref{sec:import1} illustrates how important each
modality/feature are for the classification. The classification results are
given and discussed within section \ref{sec:res}.   An alternative case study
(HABs in the Arabian Gulf) is investigated in section~\ref{sec:gulf}.  Finally,
a conclusion is given in section~\ref{sec:concs}.

\section{Review of HAB Detection Methods}
\label{ssec:Rev1}
\noindent Previous remote sensing based HAB detection methods have, in the majority of cases used spatially
isolated and single satellite sensor data samples.  Many methods have been
developed for HAB detection utilising a wide range of satellite sensors and
bands.


Many common methods of HAB detection are currently based on Chlorophyll 
concentration products, as Chl-a is in many cases, a very accurate proxy of local algal
activity. Phytoplankton is the primary water constituent \cite{Morel1977,
Morel1980} thus, Chl-a can often be accurately estimated using the water-leaving
reflectance using relationships (such as remote sensing band-ratios) for data from
sensors such as SeaWiFS, MERIS and MODIS~\cite{Matthews2011,Dierssen2010}.
The accuracy of estimating Chl-a by remote sensing sensors have aimed to be
within $\pm35\%$ in deep waters~\cite{blondeau2014review}.  However, this accuracy has not always been
found to have been met by simply using band-ratio algorithms (e.g. Moore et al.
\cite{Moore2009}).    

These simplistic methods in many cases suffer from a large quantity of false
positive detections.  The most effective updates to these methods further consider
measures of Carbon Dissolved Organic Matter (CDOM) utilising backscattering data from SeaWiFS and
MODIS~\cite{cannizzaro2008novel, tomlinson2009evaluation}. 

HAB detection using these products often use a Chlorophyll anomaly
measure that characterises the difference between today's Chl-a and a
background (often monthly or bi-monthly) average concentration~\cite{stumpf2003monitoring,
tomlinson2004evaluation, Miller2006}.  This method is also known as background
subtraction \cite{Miller2006}.

Another method of reducing the false positives associated with Chl-a HAB
detection is the backscattering ratio algorithm~\cite{cannizzaro2008novel,
tomlinson2009evaluation}.  This algorithm utilises a thresholded ratio formed
from Rrs(555) and Chl-a.   

Other optical methods have also been used such as the Spectral Shape (SS)
algorithm \cite{wynne2008relating}.  This system was proposed to discriminate
\emph{K.brevis} from other blooms creating high Chl-a values.  Alternative methods have
used both MODIS derived Fluorescence Line Height (FLH) products and locally
tuned algorithms to accommodate common inaccuracies in Chl-a estimation in
shallow coastal regions~\cite{tomlinson2009evaluation}. 

There are only a limited number of machine learning based HAB
detection/prediction systems reported in the literature.  Support Vector
Machines (SVMs) have been proposed for this application by Li et al.
\cite{liHAB} and Song et al. \cite{song2015learning}. Spatiotemporal analysis
using machine learning methods have also been proposed by Gokaraju et al.
\cite{gokaraju2012ensemble,gokaraju2011}.  Other non-machine learning methods
have been proposed for HAB monitoring, detection and prediction (e.g.
\cite{carvalho2010satellite} and those described
within~\cite{blondeau2014review}).  Machine learning techniques have also been
combined with GIS methods to produce interactive predictions~\cite{Tian2019}. 

Our work describes the definition of a unique datacube
data structure for supervised machine learning of spatiotemporal events;
specifically HAB events together with a novel machine learning architecture to
provide optimal HAB event classification and prediction performance.

\subsection{Applicable Machine Learning Methods}
\label{ssec:deep1}
\noindent Due to improved neural network models and methods combined with improvements in
computational power and the availability of extremely large ground truth datasets,
image classification performance has recently shown dramatic improvements \cite{lecun2015deep}. Deep
Convolutional Neural Networks (CNNs) are now commonly used, simple to understand
and highly effective neural networks for image classification and
characterisation~\cite{sermanet2013overfeat,girshick2014rich,krizhevsky2012imagenet,biggs2019machine}.

Recurrent Neural Networks (RNNs) \cite{bengio1994learning} have been used for state-of-the-art classification and characterisation of temporally based signals.
LSTMs (Long Short-Term Memory) are the dominant RNN form able to characterise
and model both long and short term dependencies in temporal information
\cite{hochreiter1997long,hill2018audio}.  LSTM methods have given excellent results in many
temporal characterisation problems.  However, more recently alternative methods
based on the concept of ``Attention'' have given better results in many cases
\cite{Attention}.

HAB detection requires both spatial and temporal classification.   Previous
spatiotemporal characterisation methods such as video sequence classification
\cite{donahue2015long} and multiview classification \cite{andrew2017visual} have
used and combined CNN and LSTM architectures.   HAB event characterisation is
different from these methods as the input imaging data is multimodal (in our,
case it has twelve dimensions).  We propose a novel architecture that modifies
these previous machine learning models and methods to take into account of the
multi-modal inputs.

Given that the temporal range investigated within the datacubes (see below) is
small, we
have also investigated flattening the time series sequences and utilised simple
high performance non-network classifiers as a last stage: Random Forests
\cite{RF}; Support Vector Machines \cite{SVM}; and non-temporal, fully
connected networks (Multi Layer Perceptrons: MLPs).

\section{Proposed HAB Detection System}
\label{sec:HAB1}
\noindent The proposed HAB detection system uses a supervised machine learning method.
Supervised machine learning requires a detailed ground truth dataset i.e.\
labelled positive and negative HAB events defined in time and location together
with characterising remote sensing data.

We have obtained some very large HAB event datasets from:

\begin{description}
    \item[\,\,FWC:] Florida Fish and Wildlife Conservation Commission~\cite{myfwc}
    \item[\,\,PMN:] The Phytoplankton Monitoring Network~\cite{PMN}
    \item[\,\,HAED:] The Harmful Algal Event Database~\cite{HAED}
\end{description}

\noindent We have selected the data from the Florida Fish and Wildlife Conservation
Commission (FWC) as the dataset is extremely large (of both positive and
negative HAB events) together with spanning the dates between 2001 and 2018.

We have chosen a subset of the FWC data from 2003 to June 2018 as it can be
effectively characterised by the flight times and data availability of MODIS-Aqua
and MODIS-Terra satellites and sensors.  More recent and up-to-date satellite
sensors such as Sentinel-3 have only recently become active and therefore do not
have a large amount of historical sensor data covering the date range within the
ground truth.

Only \emph{K. brevis} algae events were extracted from this dataset in order to provide a
tractable solution (\emph{K. brevis} is considered to be the most serious cause of HAB
events within the Gulf of Mexico region). 

\begin{figure}[h]
    \centering
    \includegraphics[trim={3.7cm 7.5cm 1cm 7.5cm},clip, width = 10.4cm]{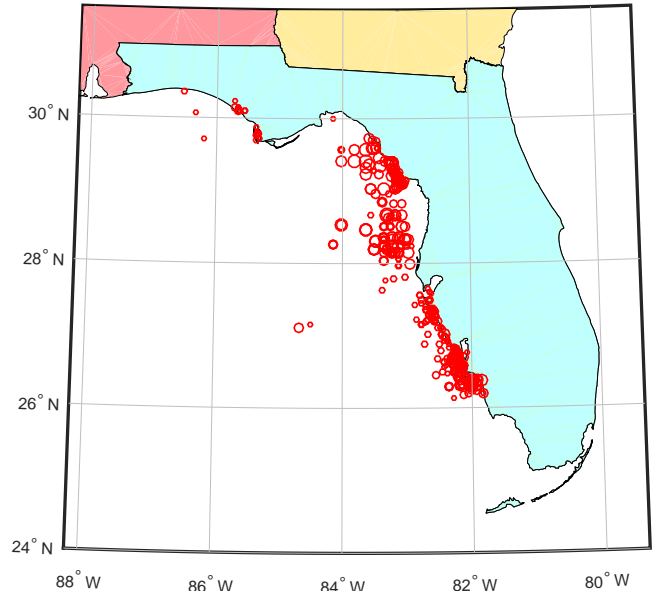}
    \caption{A selection of HAB events near Florida (2003-2018).  Circle  radius
	    reflects the log of the algae count (cells/litre) of \emph{K. brevis} HAB event}
    \label{fig:floridaHABs}
\end{figure}

In order to further reduce the size of the dataset an HAB event was considered
to have occurred when the event count algae abundance in cells/litre is in
excess of 50,000.  This is chosen as it was the threshold used in previous work
by Gokaraju et al. \cite{gokaraju2011}.  The selection of \emph{K. brevis}
events and the 50,000 threshold led to the number of positive events being 1755
(between 2003 and 2018). 1114 negative events were selected from the entire
dataset where the algae count in cells/litre were 0.  It was assumed that the
sampling positions and times for the positive and negative events were
equivalent (i.e. there was no discrimination possible between the times and
places sampled and found to be either positive or negative).  Fig.~\ref{fig:floridaHABs} shows the spatial distribution of a selection of these
positive events with the circle size reflecting the cells/litre count.

\section{Datacubes for HAB Detection}
\label{sec:Datacubes}
\noindent The most effective characterisation of HAB events for HAB event detection needs
a ``datacube'' of remote sensing data that surrounds each HAB event in time and space.

Previous datacube protocols, methods and codebases have been defined and
implemented (e.g.\ Mahecha et al.~\cite{mahecha2017emerging}).  However, these datacubes are unable
to give the required structure and/or access to remote sensing data surrounding
spatially and temporally localised events.

\begin{figure}[h!]
    \centering
    \resizebox{8.5cm}{!}{\includegraphics{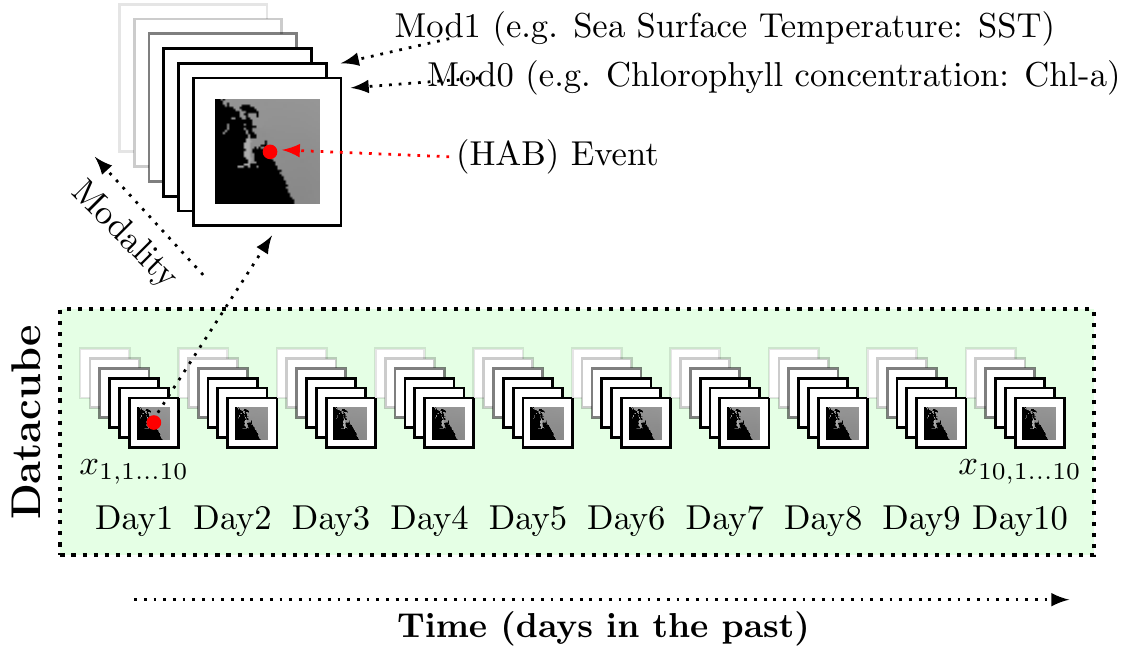}}
      \caption{Structure of a datacube used in this paper}
    \label{fig:datacube1}
\end{figure}

We have therefore developed a novel datacube definition as illustrated in
Fig.~\ref{fig:datacube1}.  Each datacube associates a range of modalities within
a spatial and temporal neighbourhood of each data point with the positive and
negative HAB ground truth database. i.e.\ there is a spatiotemporal window
defined (in metres and days) surrounding the central ground truth location (in
latitude, longitude and date).   

\noindent Extraction of remote sensing data is enabled using NASA's CMR Common
Metadata Repository (CMR) search facility~\cite{CMR}.  OB.DAAC L2 satellite file granule names are obtained given a target latitude and longitude and date of an HAB event.  These granules, in NetCDF format, are downloaded and the datacube components are extracted within the spatial and temporal neighbourhood of each HAB event.  Datacube generation is summarised in Algorithm 1 below.

\begin{algorithm}
\SetKwInOut{Input}{Input}
\SetKwInOut{Output}{Output}
\caption{Creation of ML Datacube}
\Input{Groundtruth File}
\BlankLine
\For{\emph{$\forall$ HAB events in Groundtruth File}}
{
Extract HAB event Lat, Lon, Date Window\\
\For{\emph{$\forall$ List of Modalities}}
{

Generate list of granules using NASA CMR search (within 10 days previously of HAB event
date)\\

\For{\emph{$\forall$ NetCDF Granules in Date Range}}
{
wget NetCDF Granule\\
Extract modality data in spatial window\\
Place cropped data in output Datacube\\
}
}
\Output{Datacube}
}
\label{algo_disjdecomp}
\end{algorithm}

\begin{figure}[h]
    \centering
    \resizebox{8.5cm}{!}{\includegraphics{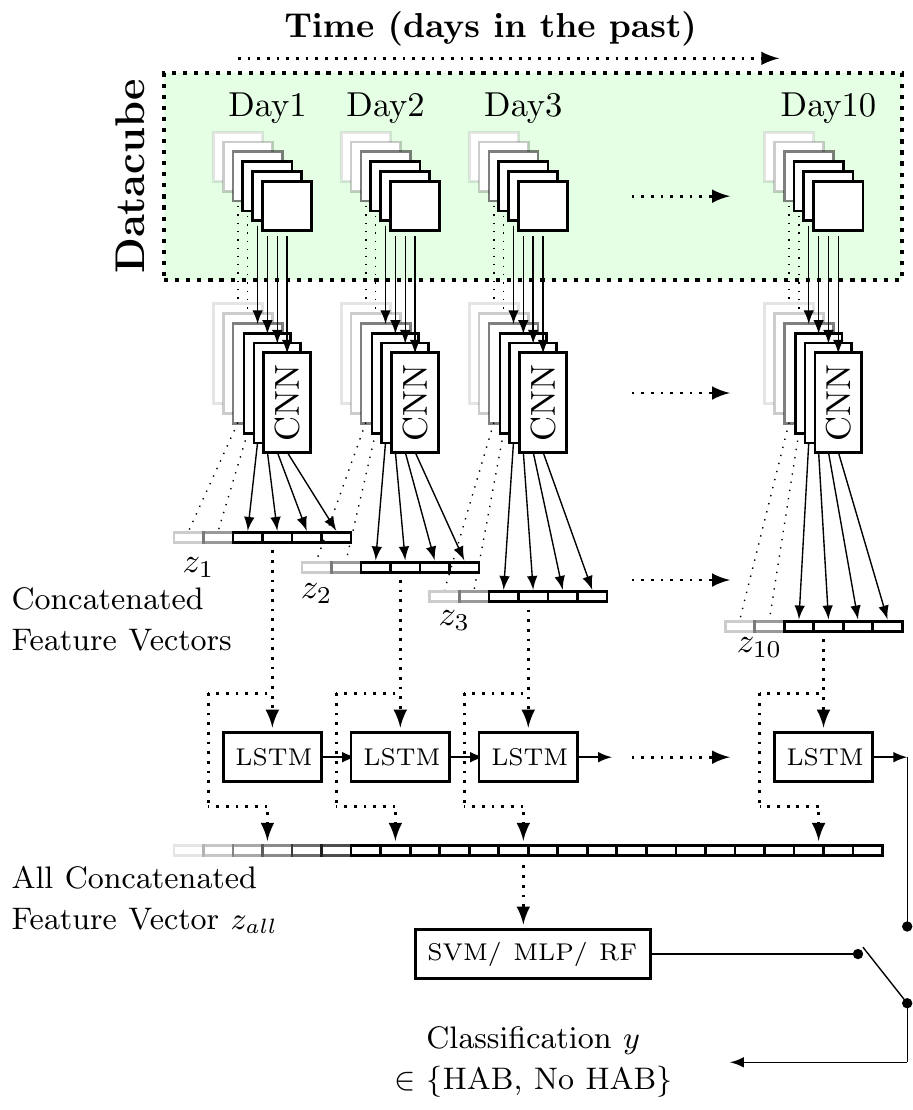}}
      \caption{Structure of HABNet Machine Learning system for datacube classification:
      CNN spatial characterisation followed by either temporal classification by
      LSTM or non-temporal classification using Multi-Layer Perceptron (MLP), Support Vector Machines (SVMs) or Random Forest (RF)
      time series classification}
    \label{fig:datacube2}
\end{figure}

\begin{figure}[h!]
    \centering
    \resizebox{8.5cm}{!}{\includegraphics{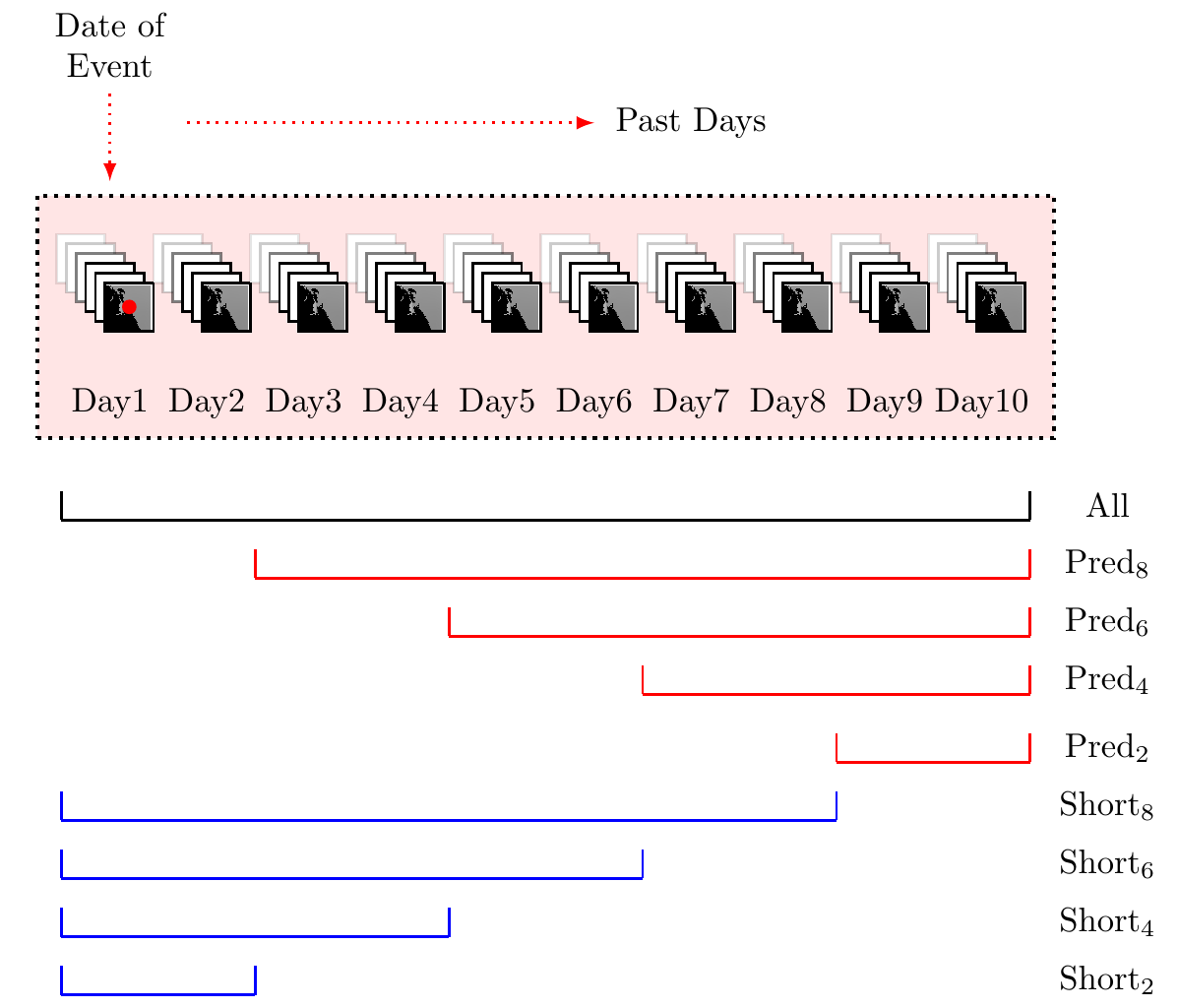}}
      \caption{Variation in temporal prediction structure using datacubes: All
      uses the entire captured datacube.  pred$_{\{2,4,6,8\}}$ sequences vary
      the number of days into the future for the training set i.e. when testing,
      the models can predict multiple days into the future.  short$_{\{2,4,6,8\}}$ sequences vary the number of days trained but do not predict into the future.  This illustrates how the number of days in the training set affects classification.}
    \label{fig:datacube40}
\end{figure}

\subsection{Selected Modalities}

\noindent The datacube architecture and the subsequent machine learning processes should
be flexible in supporting a wide range of input modalities.   However, to make
the system tractable and utilise data that is available across the whole
temporal range of the ground truth, only selected sensor data from MODIS-Aqua
and MODIS-Terra has been used.  Although using both of these satellite sensors
provides improved temporal resolution, redundancy of information has led to only
Chl-a being used from MODIS-Terra (due to a redundancy of information and the
degradation of the Terra sensor over time relative to the Aqua sensor).  Higher
level (higher than level-2) products were not considered as they lacked the
temporal and spatial resolution required for effective HAB detection and
prediction.  The level-2 products utilised are illustrated in Table
\ref{tab:feats}. 

\begin{table}[]
    \begin{center}
        \caption{List of Utilised Modalities}
	\label{tab:feats} 
\begin{tabular}{ll}
    \toprule
    \textbf{Modality}    & \textbf{Description}   \\
    \hline
    1&\textbf{Bathymetry} (GEBCO quantised from 500m grid~\cite{GEBCO})\\
    2&\textbf{MODISA Bimonthly Chl-a} (Estimated\\
&Chlorophyll concentration)\\
3&\textbf{MODISA Chl-a} (Estimated Chlorophyll concentration)\\
4&\textbf{MODISA Rrs(412)}\\
5&\textbf{MODISA Rrs(443)}\\
6&\textbf{MODISA Rrs(488)}\\
7&\textbf{MODISA Rrs(531)}\\
8&\textbf{MODISA Rrs(555)}\\
9&\textbf{MODISA PAR} (Daily Mean Photosynthetically\\ 
&Available Radiation)\\
10&\textbf{MODISA SST} (Sea surface Temperature)\\
11&\textbf{MODIST Chl-a} (Estimated Chlorophyll concentration)\\
12&\textbf{MODISA Background Anomaly} (3-2)\\
\bottomrule
\end{tabular}
\end{center}
\end{table}

Although there is possibly redundancy between the chosen bands and products such
as Chl-a, the inclusion of these bands was intended to provide fine-grain
classification and discrimination for the use of this key available data.  All
these modalities have been used by previous spatiotemporal HAB
detection methods \cite{gokaraju2011, gokaraju2012ensemble}.  Bathymetry obviously did
not vary over time, but the same modality format was used as shown where the
image samples were resampled/interpolated to be co-located as the images of the other
modalities.  Bathymetry was chosen as it has been noted that estimated
Chlorophyll concentrations are often inaccurate in shallow water.  Including
bathymetry should allow the machine learning algorithm to characterise such
variations.  A list of the utilised modalities are shown in Table~\ref{tab:feats}.

\section{Pre-Processing of Datacubes}
\label{sec:preDC}
\noindent In order to utilise the high-performance characterisation performance of
pre-trained image based CNNs, the sparse input data of the datacubes were
reprojected to a spatially consistent UTM raster image (using the standard WGS84
Ellipsoid).

The spatial representation of the input raster formats within the level-2 MODIS
based products are not spatially consistent and therefore not useful for
effective machine learning characterisation, detection and prediction.  This was
due to the capture methods and artefacts such as the ``bowtie'' effect where
horizontal or vertical lines can be repeated in the input 2D raster arrays.  The
input data\-points were therefore reprojected to locally spatially consistent UTM
reprojection. 

The projected data\-points were resampled onto a spatiotemporal grid where each
grid element was of extent (1km$\times$1km$\times$1 days) using
triangulation-based linear interpolation (the default method of the
\textsc{Matlab} \texttt{griddata} function).  The default triangulation linear
interpolation method uses a convex hull for interpolation.  The use of a convex
hull generates inaccurate resampled values where it would be better that they
were discounted.  Holes and disjoint regions are further discounted using the
\textsc{Matlab} \texttt{alphashape} function with a threshold of 0.2.   Discounted samples were
set to zero for input to the Neural Network (see below).  

Given a temporal-spatial span of 100Km and 10 days the output size of each resampled modality datacube is of dimension ($100\times100\times10$): (width$\times$height$\times$days). This structure is illustrated in Fig.~2.

\subsection{Dealing with Sparse Data}
\noindent Due to the dependence on sea surface reflectance on the majority of the chosen
modalities, a large amount of the data is missing due to cloud cover.  This is
problematic in effectively characterising any of the HAB events where there is little
data.  In order to most effectively characterise events in the ground truth
database, datacubes containing less than a threshold of data are discarded from
the training/testing process.  As the estimated Chlorophyll concentration
(Chl-a) is the most important modality within the chosen set it is used to
indicate unacceptable sparsity within dataset events.  The threshold chosen was: if
more than half of the datapoints within the Chl-a modality (averaged over the
entire temporal range) are missing, the datacube is discarded from training.

Furthermore, missing datapoints are usually indicated as NaN values in the
original OB.DAAC L2 granules.  When resampling (and the use of
\texttt{alphashape}) the original data, grid points that are not able to be
resampled are set to zero.  The use of unique flags such as zero representing
non-data is common within machine learning systems (e.g.\ Jaritz et al.\cite{jaritz2018sparse}).  As these zero values will
not be correlated with either class (HAB or non-HAB) they will not affect the
characterisation and classification of the machine learning system.  There have
been a few CNN methods proposed specifically for coping with sparse data
\cite{uhrig2017sparsity,ren2018sbnet}.  However, more recent work has indicated
that CNNs are able to learn from sparse representations directly without
explicitly changing the network structure and design~\cite{jaritz2018sparse}.

\section{Machine Learning Structure}
\label{sec:MLS}
\begin{figure}[h]
    \centering
    \includegraphics[width = 8.4cm]{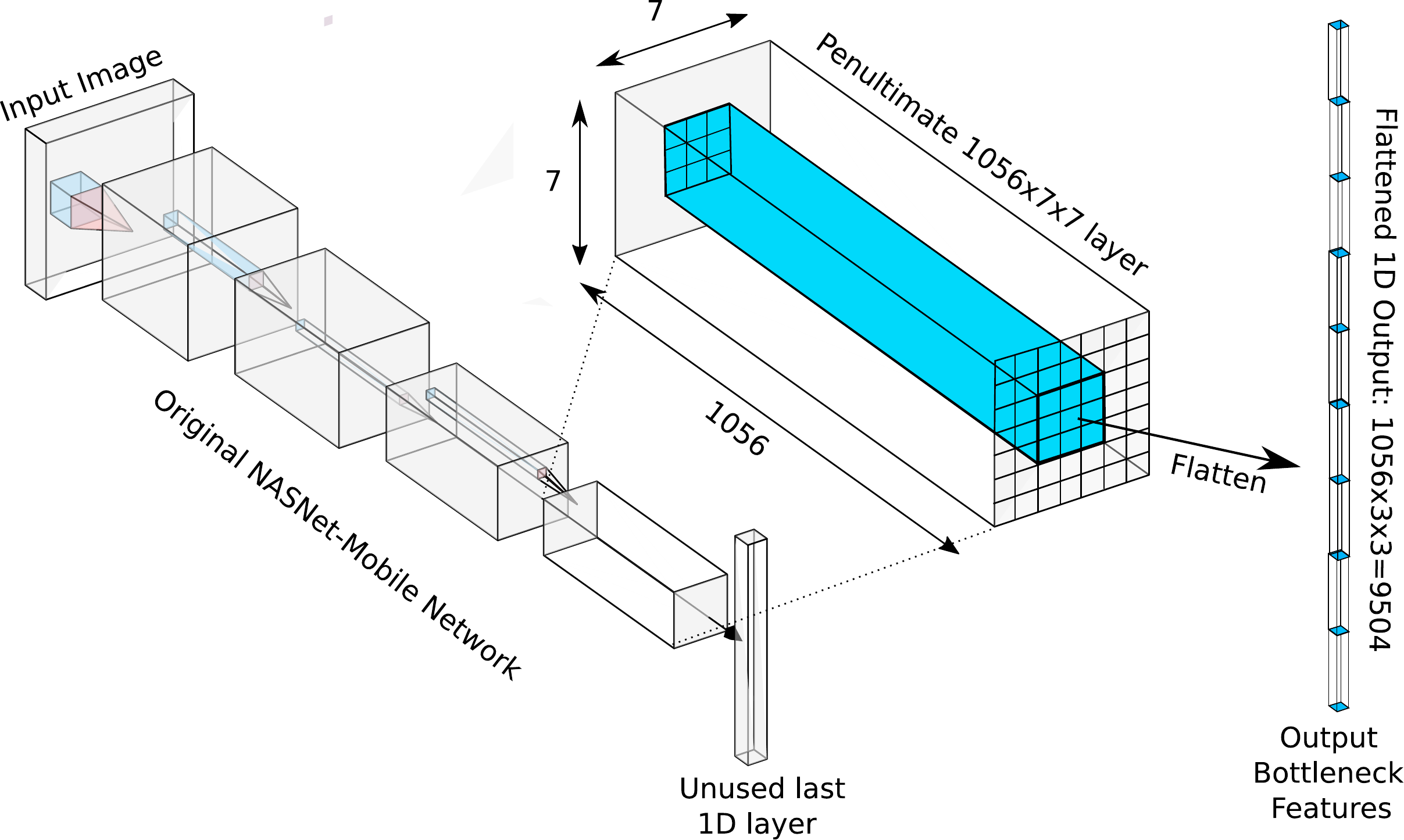}
    \caption{NASNet-Mobile extraction of translationally invariant bottleneck features.  The last NASNet layer has feature length 1056 and is has the spatial dimensions of $7\times7$.  The central region selected (to be flattened) is the central $3\times3$ regions (as illustrated).}
    \label{fig:nn3}
\end{figure}

\noindent To fully exploit the spatial and temporal discrimination information
contained within all the modalities of each datapoint, a
novel machine learning structure has been designed and implemented within this
work.  The application within this paper requires both spatial and temporal
characterisation and classification.  CNNs and LSTMs have often been combined to
provide such characterisation in applications such as video sequence
classification \cite{donahue2015long,andrew2017visual}.  However, our
application has the added complexity of multimodal 2D inputs from each quantised
time step within a temporal sequence.  We, therefore, propose the novel machine
learning structure illustrated within Fig.~\ref{fig:datacube2}.  This figure shows that a single feature vector is extracted from each single image modality
at each time step.  This is achieved through a form of transfer learning.  It
has been recognised that utilising pre-trained layer weights of existing CNNs
can provide an effective characterisation of visual features in new domains where
(as is the case in this application) there is a limited amount of training data
and computational resources. For each evaluated CNN, the final
classification layer is removed and a flattened subset of the penultimate was used as a feature
extractor (see Fig.~\ref{fig:nn3}).  These feature vectors are then concatenated for all modalities.
This concatenated feature vector is then the input to a sequenced LSTM across
all of the quantised time range (in our case for example, ten days).  A single
classification output of the LSTM is used as a binary classifier\,  $\in\{HAB, No HAB\}$.

The index of the considered time sequence is denoted $t$ where $\forall t\in
$\{$ 1,2,\dots,T$\} where in this case $T=10$.  The modality index is denoted $m$ where
$\forall m\in $\{$ 1,2,\dots,M$\} where in this case $M=12$. There are therefore
120 input images (12 modalities per each of the 10 time steps) per HAB event (each image denoted $x_{t,m}$).
The concatenated outputs $z_t$ of the CNNs are therefore created as follows.
\begin{eqnarray}
    z_t = \{ \phi(x_{t,1}), \phi(x_{t,2}),\dots, \phi(x_{t,M}) \},
\end{eqnarray}
\noindent where  $\phi(\cdot)$ is the operation of the CNN that outputs
flattened output as illustrated in Fig.~2.  

For non-temporally based classification, the concatenated outputs $z_t$ are
themselves concatenated into a single vector $z_{all}$. The LSTM temporal classification models take as input all of the concatenated outputs $z_t$ to generate the classification $y$ where $y \in \{HAB, No HAB\}$:
\begin{eqnarray}
    y = \Psi(\{ z_1, z_2,\dots, z_T\}) 
\end{eqnarray}
\noindent Where  $\Psi(\cdot)$ is the LSTM temporal classification operation that outputs the HAB/No HAB classification.

Conversely, for the non-temporally based classifiers (RF, SVM and MLPs) the
classifiers take the fully concatenated vector $z_{all}$ as input to generate
$y$: 
\begin{eqnarray}
	y = \psi(z_{all}) 
\end{eqnarray}
\noindent where  $\psi(\cdot)$ is the non-temporal classification operation (RF, SVM and MLPs) that outputs the HAB/No HAB classification.

A large number of components were tested within the architecture depicted in
Fig.~\ref{fig:datacube2}.  A variety of LSTM structures were also
tested alongside simple alternatives including MLPs (Multi-Layer Perceptrons)
and Random Forest (RF) classifiers.  Initial tests showed that NASNET:Mobile
produced the best results for the spatial CNN stage.  Table \ref{tab:results1} shows the classification results using a NASNET:Mobile CNN and a variety of temporal classification
methods.

Regularisation using $L_1$ and  $L_2$ norm conditions did not improve the results and
therefore was not used.  Standard ADAM optimisation was used with a (decaying)
learning rate of $10^{-5}$.

\subsection{List of Considered Temporal Classifiers}
\noindent Table \ref{tab:tmp} shows a list of the considered temporal classifiers (i.e.
the classifiers to take the bottleneck features generated by the CNNs and
generate binary classifications (i.e.\ $\{HAB,No HAB\}$).  All of the LSTM based
methods take the concatenated bottleneck outputs as a time series
($z_{1\dots 10}$) whereas the remaining methods take the totally
concatenated bottleneck features ($z_{all}$).

\begin{table*}[]
    \begin{center}
        \caption{List of Considered Temporal Classifiers}
	\label{tab:tmp} 
\begin{tabular}{ll}
    \toprule
    \textbf{Temporal}    & \multicolumn{1}{c}{\textbf{Description}}   \\
    \textbf{Classifier}    &  \\
    \hline
    \textbf{RF} & \textbf{Random Forest:} Standard python (\texttt{sklearn}) implementation of RF with grid
        search of best parameters using validation set\\
	\textbf{SVM} & \textbf{Support Vector Machines:} Standard python (\texttt{sklearn}) implementation of SVM with grid
        search of best parameters using validation set\\
	\textbf{MLP0} & \textbf{Multi-Layer Perceptron}: Two dense layers each with 256 nodes.  Each layer combined with batch normalisation\\
    \textbf{MLP1} & \textbf{Multi-Layer Perceptron:} Two dense layers each with 256 nodes.  Each layer combined with dropout (0.5)\\
    \textbf{MLP2} & \textbf{Multi-Layer Perceptron:} Two dense layers each with 256 nodes.  L2 normalisation\\
    \textbf{LSTM0} & \textbf{LSTM based network:} One LSTM layer and one dense layer each with 512 nodes each.  Each layer combined with batch normalisation \\&and dropout (0.5).\\
    \textbf{LSTM1} & \textbf{LSTM based network:} One LSTM layer and one dense layer each with 128 nodes.  Each layer combined with dropout (0.5).\\
    \textbf{LSTM2} & \textbf{LSTM based network:} One LSTM layer and one dense layer each with 512 nodes both returning sequences.  Each layer combined with \\&batch normalisation.\\
    \textbf{LSTM3} & \textbf{LSTM based network:} Two LSTM layers and one dense layer each with 256 nodes both returning sequences.  \\&This is flattened and then fed to fully connected layer (128 nodes).  There is dropout (0.5) and batch normalisation between each layer.\\
    \textbf{LSTM4} & \textbf{Attention based model:}  An attention layer \cite{Attention} is combined with an LSTM layer.  Dropout and batch normalisation between each layer.\\
\bottomrule
\end{tabular}
\end{center}
\end{table*}


\subsection{Translationally Variant Features}
\noindent Commonly, bottleneck features are extracted from one of the flat final 1D layers of a
CNN.  However, these features are translation invariant due to the max pooling i.e. they do not
discriminate between objects and features found in different spatial positions.
This is not what is required in this work as the HAB has been identified as
occurring specifically in the spatial centre of the input image.  In order to
make the most use of the abstractions found in the lower layer of a CNN 
the central spatial region of the penultimate layer is flattened to form
translationally variant features.  By using these flattened features without max
pooling the spatial arrangement of CNN outputs is characterised. 

\subsection{Choice of CNN and Bottleneck Features }

\noindent Many different CNN models were tested including a variety of Inception
\cite{Inception}, VGG \cite{VGG} and NASNet \cite{Nasnet} architectures.  The
choice for all the subsequent experiments was NASNetMobile as it gave the best
results and had a smaller architecture than most other models.  Large
flattened central regions will lead to excessively large bottleneck features.
For the architectures described below the choice of region size was $3\times3$.  As the
last layer of the NASNet-Mobile model has a size feature size of 1056, all of
the temporal models input bottleneck features of size $1056\times3\times3 = 9504$.  
This is illustrated in Fig.~\ref{fig:nn3}.

\section{Feature Classification Importance}
\label{sec:import1}
\begin{figure}[h]
    \centering
    \includegraphics[trim={1cm 6.5cm 1cm 8cm},clip, width = 8.4cm]{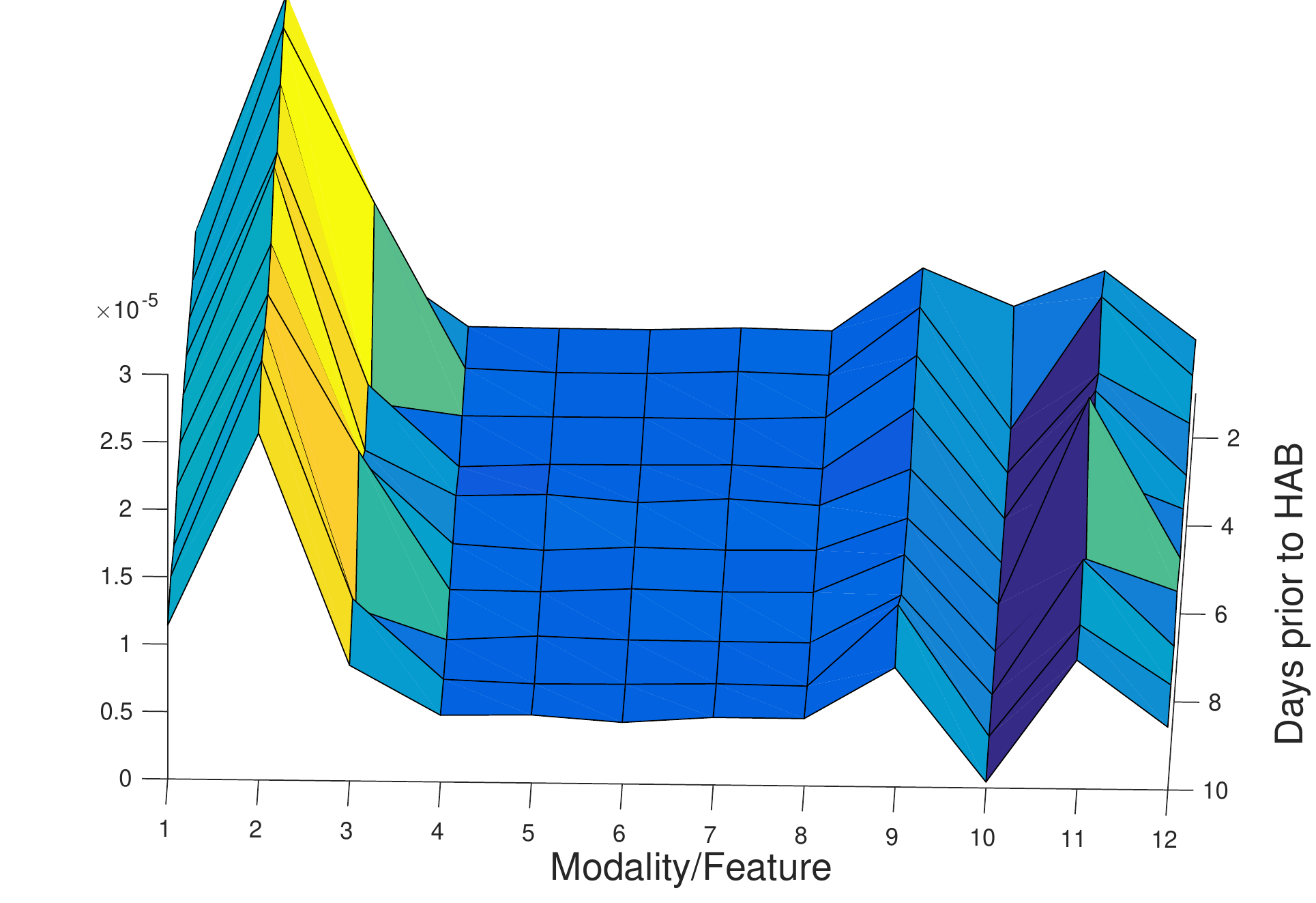}
    \caption{Feature Importances (From Random Forest Analysis~\cite{strobl2008danger}): Modality vs. Time from events.  Modality
    described in Table I.}
    \label{fig:refImports1}
\end{figure}

\begin{figure}[h]
    \centering
    \includegraphics[trim={1cm 6.5cm 1cm 8cm},clip, width = 9cm]{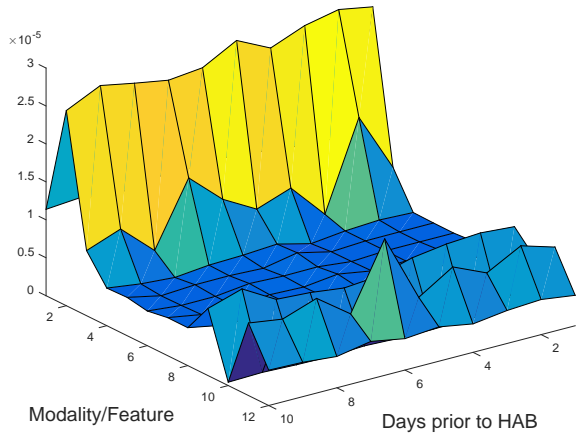}
    \caption{Feature Importances (From Random Forest Analysis~\cite{strobl2008danger}): Modality vs. Time from event. Modality
    described in Table I.}
    \label{fig:refImports2}
\end{figure}

\noindent To evaluate the classification importance of the features
given in Table~\ref{tab:feats} feature vector (input into the last temporal classifier) importance is estimated using a Random Forest classifier~\cite{RF} and its associated capability at estimating feature importances~\cite{breiman2017}.  Due to the very large feature vector length of the CNN bottleneck features, the most effective way to determine modality importance is to
extract bottleneck feature importances for the entire bottleneck feature vector
and average the importances for each combination of modality and day.  Figs.~\ref{fig:refImports1} and \ref{fig:refImports2} show the averaged importances for each modality and day.  The modality index shows those modalities labelled in Table~\ref{tab:feats} and the day indicates the number of days in the past (from the HAB event).  These figures illustrate that there is a slight decrease in importance the further in the past the features are.  Additionally, it is apparent that the most important features are those indexed\,\, \{1,2,3,9,11\}.  This indicates that the individual $Rrs$ reflectances\,\,  \{4,5,6,7,8\}, SST \{10\} and Background anomaly \{12\} features are relatively unimportant (in terms of classification). \footnote{these ``unimportant modalities'' are therefore omitted in a subset of subsequent experiments in order to reduce the concatenated feature length and therefore the computational load and memory requirements}

\section{Results}
\label{sec:res}
\noindent The most effective way to evaluate the performance of a classifier is using nested cross validation \cite{goodfellow2016deep}.  Nested cross validation utilises two nested cross validation stages.  The outer cross validation iteratively splits the whole dataset (2869 datapoints) into 5 folds in our case (each fold having separate training and testing subsets).  For each of the outer folds the outer training set is further split into training and validation sets.  The inner validation set is used to validate and optimise model and parameter choice.   The results shown in Tables \ref{tab:results1}, \ref{tab:results2} and \ref{tab:results3} show the average and standard deviation across all 5 outer folds. 

The best results were obtained using the LSTM3 model combined with a
NASNET:Mobile CNN (mean accuracy 91\% and Kappa coefficient 0.81).  Although not
comparable in terms of dataset location, time and size these results are
relative improvements to correct classification rates of 83\%
\cite{tomlinson2004evaluation} and 75\% \cite{stumpf2003monitoring} for standard
Chlorophyll anomaly detection methods.  These compared methods both used
extremely small datasets compared to our work.  Furthermore, the results show
that the reduction in the number of features does not significantly decrease the
performance of the classifiers.  Additionally, these results show that
considering the outputs of the bottleneck features as a time series and
analysing them with an appropriate time series analysis tool such as an LSTM or
attention based network does not give significant improvements in performance
compared to non time series classification tools such as SVMs.  This is
considered to be because time series of length 10 (temporal data points) are difficult to
evaluate using such temporal analysis tools.

\subsection{Conventional Comparative Methods}

\noindent Comparable methods of HAB detection include Spectral Shape (SS), thresholded
backscattering ratio~\cite{cannizzaro2008novel, tomlinson2009evaluation} and
thresholded Chl-a anomaly \cite{stumpf2003monitoring, tomlinson2004evaluation,
Miller2006}. These three methods are used to compare the performance with our
developed methods.

The spectral shape gives a measure of the spectral shape centred on a specific
band. It is based on a simple measure combining the Remote Sensing Reflectance
(Rrs) of neighbouring reflectance bands:
\begin{align}
SS(\lambda)=& nLw(\lambda)-nLw(\lambda^-)\nonumber \\
&-(nLw(\lambda^+)-nLw(\lambda^-)) \times \left(\frac{\lambda-\lambda^-}{\lambda^+-\lambda^-}\right),	
\label{SS}
\end{align}

\noindent where $nLw$ is water leaving radiance at wavelength in nanometers
($nLw$ is linearly related band reflectances: $Rrs$). Equation (\ref{SS}) is a
second derivate measure centered on wavelength $\lambda$.  The results in Table
V for spectral shape are for MODISA [$SS(488)$] as used by Stumpf et
al.~\cite{stumpf2010adjustment} and close to $SS(490)$ used for SeaWifs sources by
Tomlinson et al. for \emph{K. brevis} in the Florida region
\cite{tomlinson2009evaluation}.

Another method of reducing the false positives associated with Chl-a HAB
detection is the backscattering ratio algorithm~\cite{cannizzaro2008novel,
tomlinson2009evaluation}.  This algorithm utilises a thresholded ratio formed
from Rrs(555) and Chl-a.   For SeaWiFS bands the backscatter ratio is given as:
\begin{align}
	bp_{ratio}=\frac{b_{bp}\left(555\right)}{b_{bp}\left(555\right)_{Morel}},
	\label{bpRat}
\end{align}
\noindent where $b_{bp}\left(555\right) = -0.00182+2.058\times Rrs(555)$\cite{Cannizzaro2008,Roesler2002} and
$b_{bp}\left(555\right)_{Morel}=0.3\times Chl-a^{0.62} \times (0.002+0.02\times
(0.5-0.25\times log_{10}(Chl-a)))$ \cite{morel1988optical}.

Detection of HABs utilise this ratio together with a conventional threshold of
1.0 \cite{Cannizzaro2008} and a threhsold of 2.0 for non K. brevis based HAB
detection~\cite{tomlinson2009evaluation}.  Table \ref{tab:conv} shows the
results on the entire dataset of 2869 datapoints.  These values are very low
compared to our developed method.  

Although difficult to compare directly, Gokaraju et al. have developed limited
spatiotemporal methods of HAB detection using SVMs and Neural Networks
\cite{gokaraju2011, gokaraju2012ensemble}. These
pieces of work both use ground truth in Florida but only a very small data set
(less than 30 datapoints for MODIS based estimation).  The best kappa
classification for the MODIS-A data was 0.65.

\subsection{Results: Discussion}
\noindent All of the results for the developed methods for HAB classification shown in Table III gave
significantly better results than the conventional methods shown in Table V.
It is assumed that adopting a two stage machine learning based approach
is able to much more effectively characterise spatial and temporal
discriminating information for HAB classification relative to these
conventional methods.   Further observations from the results include:

\begin{itemize}
	\item Larger spatial areas included in the output of the CNN bottelneck features improve results (i.e.\ going from $1\times1$ to $3\times3$ improves results.)
	\item The selection of a smaller number of the most important features
		does not decrease results (i.e.\ going from all features
		\{1\dots12\} to\,\, \{1,2,3,9,11\} does not decrease classification
		performance results).
	\item Temporally based networks such as LSTMs do not provide
		significantly better results than ``flat'' solutions such as MLPs or SVMs.  This was assumed to be because time series with only 10 datapoints are not long enough for LSTMs to effectively characterise temporal variations over such a short amount of time. 
\end{itemize}

\begin{table}[]
    \begin{center}

        \caption{Classification Accuracy Results: Florida}
        	    \label{tab:results1}
\begin{tabular}{lllll}
	    \toprule
    \textbf{Features}    & \textbf{Temporal}    & \textbf{Accuracy} & \textbf{F1} & \textbf{Kappa (K)}  \\
    \textbf{(Selected from} & \textbf{Classifier} & Mean$\pm$SD&Mean$\pm$SD&Mean$\pm$SD\\
    \textbf{NASNet:Mobile)} & \textbf{ } & &&\\
    \hline
    \{1 to 12\}&      LSTM0 & 0.89$\pm$0.02&0.86$\pm$0.03&0.77$\pm$0.05\\
    \{1,2,3,9,11\}&   LSTM0 & 0.88$\pm$0.04&0.84$\pm$0.04&0.74$\pm$0.08\\
    \{1,2,3,9,11\}&   LSTM1 & 0.89$\pm$0.03&0.86$\pm$0.04&0.78$\pm$0.06\\
    \{1,2,3,9,11\}&   LSTM2 & 0.88$\pm$0.04&0.84$\pm$0.05&0.74$\pm$0.08\\
    \{1,2,3,9,11\}&   LSTM3 & \textbf{0.91}$\pm$0.02&\textbf{0.88}$\pm$0.03&\textbf{0.81}$\pm$0.05\\
    \{1,2,3,9,11\}&   LSTM4 & 0.91$\pm$0.03&0.88$\pm$0.05&0.80$\pm$0.05\\
    \{1,2,3,9,11\}&   RF &    0.87$\pm$0.04&0.82$\pm$0.06&0.72$\pm$0.08\\
    \{1,2,3,9,11\}&   SVM &   0.90$\pm$0.03&0.88$\pm$0.04&0.80$\pm$0.06\\
    \{1,2,3,9,11\}&   MLP0 & 0.89$\pm$0.02&0.86$\pm$0.02&0.77$\pm$0.04\\
    \{1,2,3,9,11\}&   MLP1 & 0.73$\pm$0.21&0.46$\pm$0.75&0.37$\pm$0.61\\
    \{1,2,3,9,11\}&   MLP2 & 0.89$\pm$0.03&0.86$\pm$0.04&0.77$\pm$0.06\\
    \bottomrule
\end{tabular}
\end{center}
\end{table}

\begin{table}[]
    \begin{center}
        \caption{Classification Accuracy Results: Florida Results for Varying Temporal
	Ranges (Ranges shown in Fig. 4)}
        \label{tab:results2}
\begin{tabular}{llll}
    \toprule
    \textbf{Temporal}    & \textbf{Accuracy} & \textbf{F1} & \textbf{Kappa (K)}  \\
    \textbf{Range} & Mean$\pm$SD&Mean$\pm$SD&Mean$\pm$SD\\
    \hline
    All(LSTM3) & \textbf{0.91}$\pm$0.02&\textbf{0.88}$\pm$0.03&\textbf{0.81}$\pm$0.05\\
    pred$_8$ & 0.90$\pm$0.02&0.87$\pm$0.02&0.79$\pm$0.04\\
    pred$_6$ & 0.90$\pm$0.01&0.87$\pm$0.01&0.79$\pm$0.02\\
    pred$_4$ & 0.89$\pm$0.03&0.86$\pm$0.04&0.76$\pm$0.07\\
    pred$_2$ & 0.88$\pm$0.02&0.85$\pm$0.03&0.74$\pm$0.05\\
    short$_2$ & 0.87$\pm$0.01&0.84$\pm$0.03&0.73$\pm$0.04\\
    short$_4$ & 0.88$\pm$0.05&0.85$\pm$0.06&0.75$\pm$0.10\\
    short$_6$ & 0.89$\pm$0.02&0.86$\pm$0.04&0.76$\pm$0.05\\
    short$_8$ & 0.90$\pm$0.01&0.87$\pm$0.03&0.79$\pm$0.03\\
    \bottomrule
\end{tabular}
\end{center}
\end{table}

\begin{table}[]
    \begin{center}
        \caption{Classification using conventional thresholds: SS(488) defined
		as \ref{SS},  $Bp_{Ratio}$ defined as \ref{bpRat} and $ChlaAnom$ is defined as
		feature 12 in Table~\ref{tab:feats} (i.e.\ feature 3 - feature 2
	in the same table).}
       \label{tab:conv}
\begin{tabular}{llll}
    \toprule
    \textbf{Method}    & \textbf{Accuracy} & \textbf{F1} & \textbf{Kappa (K)}  \\
    \hline
    $SS(488)$ $<0.0$ & 0.59&0.13&0.00\\
    $Bp_{Ratio}$ $<1.0$ & 0.61&0.44&0.15\\
    $Bp_{Ratio}$ $<2.0$ & 0.55&0.53&0.13\\
    $Chla\_Anom$ $>1.0$ & 0.58&0.29&0.04\\
    $Chla\_Anom$ $>10.0$ & 0.62&0.23&0.08\\
    $Chla\_Anom$ $>100.0$ & 0.61&0.10&0.03\\
    \bottomrule
\end{tabular}
\end{center}
\end{table}

\section{Alternative Case Study: HABs in the Arabian Gulf}
\label{sec:gulf}
\noindent As an alternative case study region, the Arabian Gulf was chosen at it has
significantly different environmental factors and the availability of ground
truth data.  HABs occur regularly within the Arabian Gulf with serious outbreaks
having happened most years over the last few decades
\cite{rao1998phytoplankton,heil2001first,glibert2002fish,moradi2012red}.

Over 38 types of algal taxa have been identified in the Arabian Gulf
\cite{zhao2014monitoring}.  A very serious outbreak in 2008 affected over 1200
km of coastline while destroying thousands of tons of fish and marine life.
Such serious HAB events can do considerable damage to local aquaculture and can
potentially shutdown vital local desalination plants (a major source of local
potable water \cite{berktay2011environmental}).

A number of local research projects into HAB monitoring and prediction have been
undergone in the last ten years \cite{zhao2014monitoring,Zhao2016}.  These
methods have focused on remote sensing data such as MODIS-A and MERIS.  However,
these works have not generated quantitative results in the detection and/or
prediction of a database of HAB events within the Gulf.  They instead have used
HAB indicators focused on optical measures such as the modified Fluorescence
Line Height (mFLH) and enhanced Red–Green–Blue (eRGB) measures together with
flow models such as HYCOM \cite{zhao2014exploring}.

\subsection{Ground Truth}

\begin{figure}[h]
    \centering
    \includegraphics[trim={2cm 7cm 2cm 7cm},clip, width = 8.4cm]{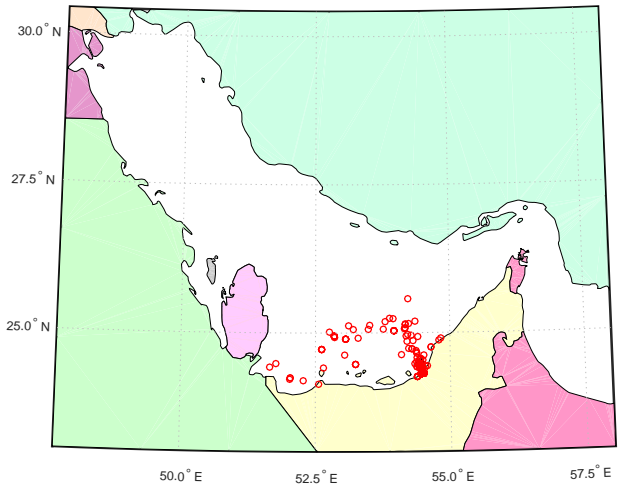}
    \caption{A selection of positive HAB events recorded in the Arabian Gulf off the coast of the UAE between
    2003 and 2018.}
    \label{fig:ArabianG1}
\end{figure}

\noindent A considerably smaller set of ground truth HAB events (compared to that obtained
for the Florida area by the FWC) have been obtained from the Environment
Agency-Abu Dhabi (EAD) between the dates of 2002 and 2018 (covered by the flight
times of MODIS).  This ground truth dataset contains 249 positive events from
multiple species (from generic labels such as Cyanobacteria, specific HAB
detection species such as Cochlodinium and multi-species detections).  Alongside
the 249 positive events 374 negative events were generated that were distinct
(in time and space from the positive events) within the same Arabian gulf region
(off the coast of Abu Dhabi: see Fig.~\ref{fig:ArabianG1}).  These detections
were not accompanied by concentrations (in cells/litre).  This means it was
implicitly assumed they were ``significant events''.  However, this does render
it difficult to make this case study analysis comparable to the Florida case
above (as within the Florida case there was an explicit threshold of
concentrations that constituted an HAB ``event''). 

\subsection{Results}
\noindent As per the mechanisms described in section \ref{sec:Datacubes}, datacubes were
extracted for all the spatial locations given by the positive and negative events.
The datacubes were comprised of the same list of modalities listed in Table
\ref{tab:feats}.  However, (due to the same analysis given in section \ref{sec:import1}) the
modalities used for classification were actually a subset of those included in
the datacubes (as shown in Table~\ref{tab:results3}). 

The results are shown in Table~\ref{tab:results3} giving classification
accuracy, F1 and Kappa metrics for the given temporal classifiers.  As with the
previous work in Florida, the temporal CNN analysis (to produce bottleneck
features) was the NASNet:Mobile CNN producing flattened $3\times3$
translationally variant features.  The temporal classification stages (LSTM0,
LSTM1 etc.) are defined as shown in Table \ref{tab:tmp}.

This table shows that SVMs in this case give the best classification results (in
terms of classifier accuracy, F1 and the Kappa coefficient).

\begin{table}[]
    \begin{center}
        \caption{Classification Accuracy Results: Arabian Gulf}
	\label{tab:results3}
\begin{tabular}{lllll}
    \textbf{Features Selected}    & \textbf{Temporal}    & \textbf{Accuracy} & \textbf{F1} & \textbf{Kappa}  \\
    \textbf{(from} & \textbf{Classifier} & Mean$\pm$SD&Mean$\pm$SD&Mean$\pm$SD\\
    \textbf{NASNet:Mobile} & \textbf{ } & &&\\
    \hline
    \{1,2,3,9,11\} &   LSTM0 & 0.91$\pm$0.05&0.88$\pm$0.06&0.82$\pm$0.10\\
    \{1,2,3,9,11\} &   LSTM1 & 0.88$\pm$0.08&0.81$\pm$0.14&0.73$\pm$0.19\\
    \{1,2,3,9,11\} &   LSTM2 & 0.86$\pm$0.06&0.82$\pm$0.08&0.70$\pm$0.12\\
    \{1,2,3,9,11\} &   LSTM3 & 0.91$\pm$0.09&0.88$\pm$0.11&0.80$\pm$0.18\\
    \{1,2,3,9,11\} &   LSTM4 & 0.91$\pm$0.06&0.87$\pm$0.13&0.80$\pm$0.16\\
    \{1,2,3,9,11\} &   RF &    0.91$\pm$0.06&0.88$\pm$0.08&0.81$\pm$0.12\\
    \{1,2,3,9,11\} &   SVM &   \textbf{0.93}$\pm$0.08&\textbf{0.91}$\pm$0.09&\textbf{0.85}$\pm$0.16\\
    \{1,2,3,9,11\} &   MLP0 & 0.91$\pm$0.07&0.88$\pm$0.11&0.81$\pm$0.16\\
    \{1,2,3,9,11\} &   MLP1 & 0.71$\pm$0.23&0.31$\pm$0.77&0.27$\pm$0.66\\
    \{1,2,3,9,11\} &   MLP2 & 0.90$\pm$0.09&0.85$\pm$0.16&0.78$\pm$0.21\\
\end{tabular}
\end{center}
\end{table}

\section{Conclusion}
\label{sec:concs}
\noindent HABNet is the first machine learning architecture to use remote sensing based datacubes defined
specifically for the classification (using machine learning methods) of
temporally and spatially isolated events such as HAB events.  A very large
database of positive and negative HAB events was utilised over the last two
decades off the coast of Florida.  Seadas tools combined with NASA's CMR
web-based enquiry method were used to populate a ground truth database of
datacubes (one per data point in the ground truth database).  12 modalities were
chosen including estimated sea surface temperature, Chlorophyll concentrations,
reflectance bands (from MODIS sensors) and bathymetry.  A combined CNN/LSTM
spatiotemporal classification system was implemented to classify and
discriminate between HAB and non-HAB events.  Using a small NASNET-mobile CNN
with an LSTM temporal stage a classification accuracy and Kappa coefficient of
91\%  and 0.81 were achieved respectively.  This is a significant improvement
compared to results generated from historical methods (e.g. Chl-a anomaly:
Maximum Kappa = 0.08) and other reported state of the art classification methods
(spatiotemporal classification method using for a very small dataset: Maximum
Kappa = 0.65).  Our results represent a significant correct classification rate
(and Kappa coefficient) given that the number of datapoints is an order of
magnitude greater than any previous study. In the future, targeted integration
of supplementary modalities and optimisation of machine learning methods and
structures are anticipated to lead improved classification rates.

Furthermore, our study shows that the datacube method is able to effectively
predict HABs up to 8 days in the future without significant degradation of
classification accuracy.

An alternative case study was investigated for multi-species HAB ground truth
events within the Arabian
Gulf.  Good results were also obtained from this study given a maximum
classification rate of 93\% and a Kappa coefficient of 0.85 (using a
NASNet-Mobile CNN and an SVM temporal stage).

A transfer learning
method to use the present work for characterisation transferred to the use of
more up to data sensors such as Sentinel-3 would be an essential follow-up
project.  Furthermore, a study of the effect of spatial and temporal resolution
on classification performance would also be key for subsequent studies.

\section*{Acknowledgment}
\noindent The authors would like to thank the British Council for funding: British Council Award ref 279334808.  The FWC for the ground truth data in Florida and EAD (UAE) for the ground truth data in the
Arabian Gulf.
\bibliographystyle{IEEEtran}
\bibliography{IEEEabrv,refs}
\end{document}